\documentclass[]{modified_llncs}
\usepackage{graphicx}
\usepackage{amsmath,amssymb} 
\usepackage{authblk}

\usepackage{multicol}
\usepackage{multirow}
\usepackage{float} 
\usepackage{hyperref}
\hypersetup{
    colorlinks=true,
    linkcolor=blue,
    filecolor=magenta,      
    urlcolor=cyan,
    pdftitle={Overleaf Example},
    pdfpagemode=FullScreen,
    }

\usepackage{algorithm} 
\usepackage{algpseudocode} 

\usepackage{color}

\begin{document}
\pagestyle{headings}
\mainmatter

\title{Unified Object Detector for Different Modalities based on Vision Transformers} 


\author[1]{Xiaoke Shen}
\author[2,3]{Ioannis Stamos}
\affil[1]{Mobi Systems, Inc, 48 Grove St, Somerville, 02144, MA, USA; jimmy@takemobi.com}
\affil[2]{Hunter College, CUNY \\
New York City, USA}
\affil[3]{The Graduate Center, CUNY \\
New York City, USA}
\affil[ ]{\textit {istamos@hunter.cuny.edu}}
\date{}
\maketitle

\abstract{Traditional systems typically require different models for processing different modalities, such as one model for RGB images and another for depth images. Recent research has demonstrated that a single model for one modality can be adapted for another using cross-modality transfer learning. In this paper, we extend this approach by combining cross/inter-modality transfer learning with a vision transformer to develop a unified detector that achieves superior performance across diverse modalities. Our research envisions an application scenario for robotics, where the unified system seamlessly switches between RGB cameras and depth sensors in varying lighting conditions. Importantly, the system requires no model architecture or weight updates to enable this smooth transition. Specifically, the system uses the depth sensor during low-lighting conditions (night time) and both the RGB camera and depth sensor or RGB caemra only in well-lit environments. We evaluate our unified model on the SUN RGB-D dataset, and demonstrate that it achieves similar or better performance in terms of mAP50 compared to state-of-the-art methods in the SUNRGBD16 category, and comparable performance in point cloud only mode. We also introduce a novel inter-modality mixing method that enables our model to achieve significantly better results than previous methods. We provide our code, including training/inference logs and model checkpoints, to facilitate reproducibility and further research. \url{https://github.com/liketheflower/UODDM}}

\section{Introduction}
\label{sec:intro}
Advances in computer vision and artificial intelligence have enabled the development of increasingly sophisticated robotic applications that enhance human lives. Autonomous vehicles, for instance, can transport individuals to their destination without the need for a human driver/operator, while autonomous mobile robots operating in warehouses can assist in order preparation. However, many robotic systems rely on multiple sensors, such as cameras and 3D sensors (LiDAR or depth), and not all sensors are equally effective in all scenarios. For instance, camera sensors may perform poorly in low-light conditions without supplementary lighting. Thus, the ability to operate in low-light conditions can significantly reduce electricity usage and promote environmentally friendly robot design.\\
\begin{figure}[t]
\begin{center}
   \includegraphics[width=1.0\linewidth]{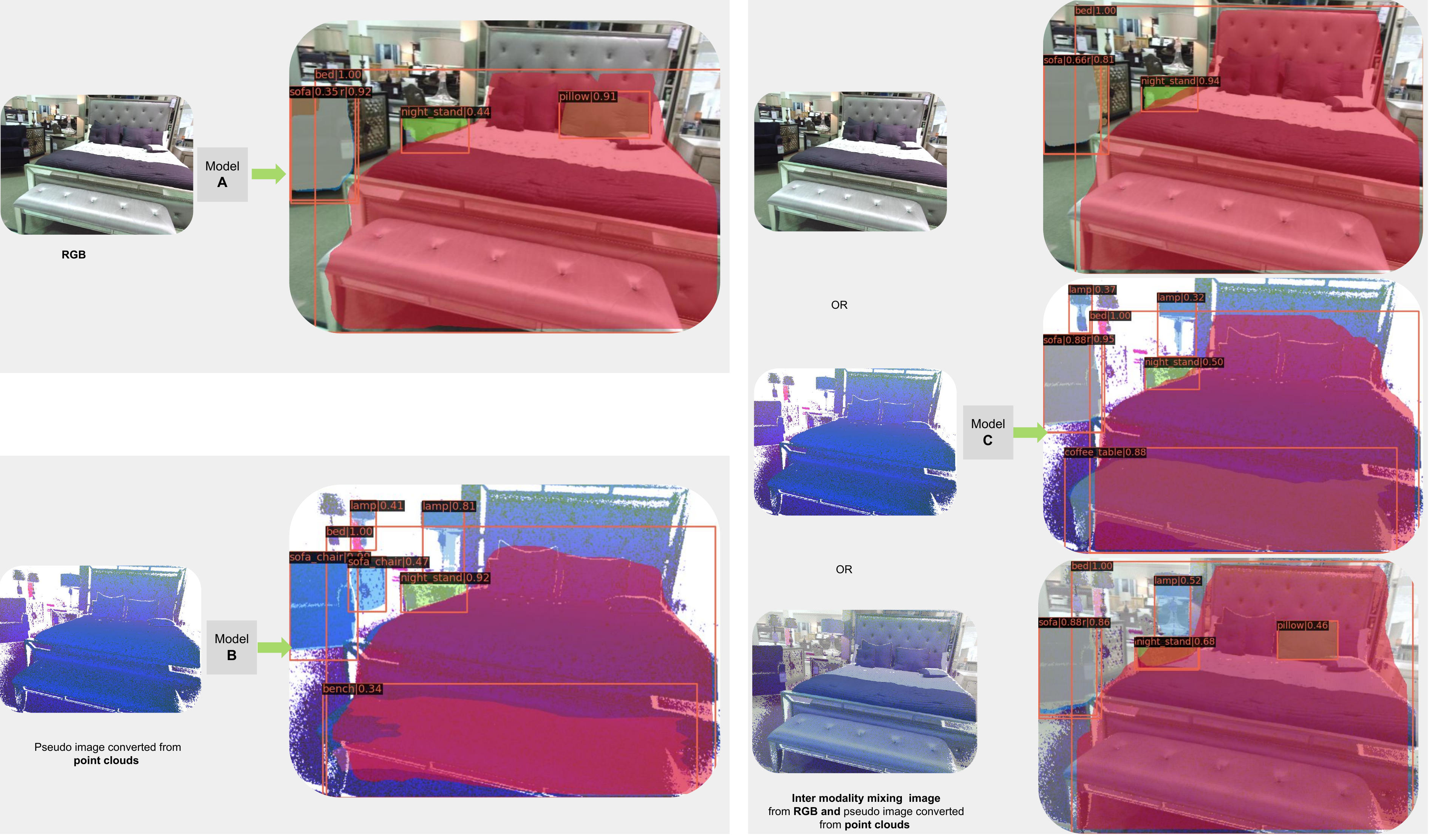}
\end{center}
    \caption{Model A exclusively processes RGB images, with the visualization generated solely from the RGB-trained model presented in this study. Model B operates on pseudo images converted from point clouds, and the visualization is derived from the simCrossTrans\cite{simCrossTrans_cite} approach, which trains on these images. Model C is capable of processing RGB images, pseudo images converted from point clouds, or a combination of both. The visualization is based on UODDM with a Swin-T \cite{liu2021Swin} backbone network.}
\label{fig:main}
\end{figure}

The high accuracy achieved by camera-based vision systems in 2D detection owes much to the efficacy of ConvNets\cite{10.1162/neco.1989.1.4.541} and Transformer\cite{transformer_cite} based feature extractors. Concurrently, simCrossTrans\cite{simCrossTrans_cite} proposes that by converting 3D sensor data into pseudo images and applying cross-modality transfer learning, a 2D object detection system using identical networks as those used for RGB images can produce commendable results. This development prompts a natural question: can we further enhance performance by training a unified network with both RGB and 3D data, adopting an identical architecture and weights throughout? The proposed unified network accepts three types of sensor data, namely, 1) RGB images, 2) pseudo images converted from 3D sensors, and 3) both RGB images and pseudo images converted from 3D sensors. If a unified network can match or exceed the detection performance of separate networks, each optimized for a particular modality, it would make feasible the use of an eco-friendly system operating under natural lighting conditions during the day and without any extra lighting at night. This article examines the potential of such a system. Building on simCrossTrans\cite{simCrossTrans_cite} that demonstrates the superior performance of a Vision Transformer-based network over ConvNets-based networks, our study concentrates exclusively on the Vision Transformer network.\\

\begin{figure}[H]
\begin{center}
   \includegraphics[width=1.0\linewidth]{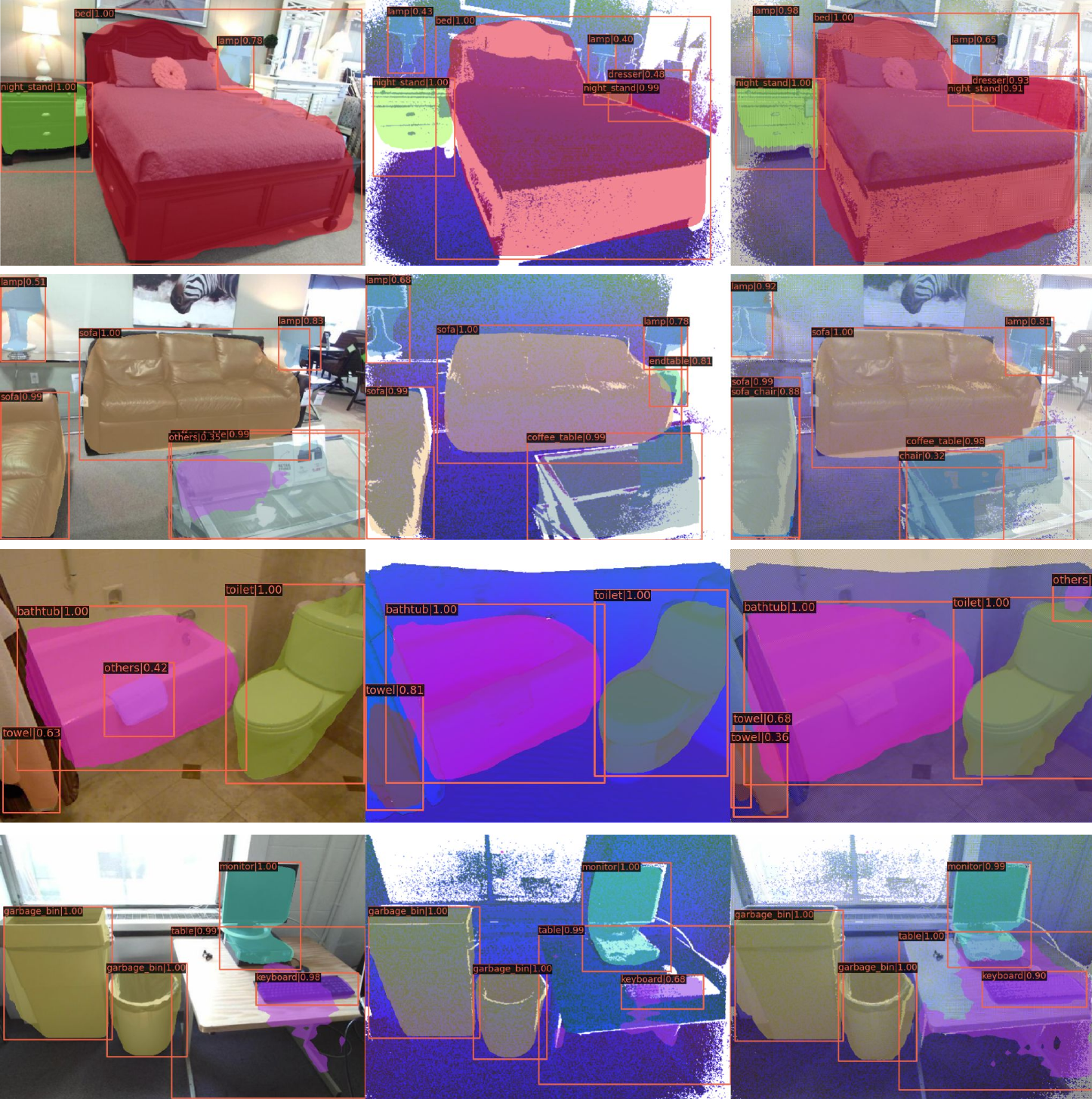}
\end{center}
    \caption{2D detection results of UODDM, using the SUN RGB-D validation dataset. It showcases four examples of 2D detection visualization, with the left column showing RGB images, the middle column displaying pseudo images converted from point clouds, and the right column illustrating inter-modality mixing of RGB images and pseudo images converted from point clouds. The backbone network used in this study is Swin-T. More demo can be found \url{https://youtu.be/PuRQCLLSUDI}}
\label{fig:example}
\end{figure}
In summary, this article aims to address the following research questions:\\
\begin{enumerate}
    \item Can a unified model achieve comparable or superior performance in processing both RGB images and pseudo images converted from point clouds?
    \item If a unified model that processes both RGB and pseudo images is feasible, can the RGB and pseudo images be further fused to enhance the model's ability to process both RGB and point cloud data?
\end{enumerate}

We conducted experiments that resulted in insightful observations and achieved state-of-the-art performance in 2D object detection. Our proposed unified model, named the Unified Object Detector for Different Modalities (UODDM), is capable of processing various types of images, including RGB images, pseudo images converted from point clouds, and inter-modality mixing of RGB image and pseudo images converted from point clouds. Figure \ref{fig:main} illustrates the differences between our model and other works. Furthermore, the performance comparison of different methods can be found in Table \ref{2d_res}. Visualizations of UODDM outputs are presented in Figure \ref{fig:example}.\\
The key contributions of our work can be summarized as follows:\\
\begin{enumerate}
\item We propose two inter-modality mixing methods which can combine the data from different modalities to further feed to our unified model.
\item We propose a unified model which can process any of the following images: RGB images, pseudo images converted from point clouds or inter-modality mixing of RGB image and pseudo images converted from point clouds. This unified model achieves similar performance to RGB only model and point cloud only model. Meanwhile, by using the inter-modality mixing data as input, our model can achieve a significantly better 2D detection performance.
\item We open source our code, training/testing logs and model checkpoints.
\end{enumerate}
\section{Related Work}
\textbf {Projecting 3D sensor data to 2D Pseudo Images: } There are different ways to project 3D data to 2D features. HHA was proposed in \cite{GuptaGAM14} where the depth image is encoded with three channels: Horizontal disparity, Height above ground, and the Angle of each pixel’s local surface normal with gravity direction. The signed angle feature described in\cite{6375012} measures the elevation of the vector formed by two consecutive points and indicates the convexity or concavity of three consecutive points. Input features converted from depth images of normalized depth(D), normalized relative height(H), angle with up-axis(A), signed angle(S), and missing mask(M) were used in \cite{7785116}. DHS images are used in \cite{Shen_2020_WACV,Shen_2020_v2}. \\

\textbf{Object Detection Based on RGB images or Pseudo images from point cloud by Vision Transformers: }
Object detection approaches can be summarized as two-stage frameworks (proposal and detection stages) and one-stage frameworks (proposal and detection in parallel). Generally speaking, two-stage methods such as R-CNN \cite{DBLP:journals/corr/GirshickDDM13}, Fast RCNN \cite{DBLP:conf/iccv/Girshick15}, Faster RCNN \cite{DBLP:conf/nips/RenHGS15}, FPN \cite{Lin_2017_CVPR} and mask R-CNN \cite{maskrcnn} can achieve a better detection performance while one-stage systems such as YOLO\cite{DBLP:journals/corr/RedmonDGF15}, YOLO9000\cite{DBLP:journals/corr/RedmonF16}, RetinaNet \cite{DBLP:journals/corr/abs-1708-02002} are faster at the cost of reduced accuracy. For deep learning based systems, as the size of the network is increased, larger datasets are required. Labeled datasets such as PASCAL VOC dataset \cite{pascal-voc-2012} and COCO (Common Objects in Context) \cite{DBLP:journals/corr/LinMBHPRDZ14} have played important roles in the continuous improvement of 2D detection systems. Most systems introduced here are based on ConvNets. Nice reviews of 2D detection systems can be found in \cite{DBLP:journals/corr/abs-1905-12683}. When replacing the backbone network from ConvNets to Vision Transformers, the systems will be adopted to Vision Transformers backbone based object detection systems. The most successful systems are Swin-transformer\cite{liu2021Swin} and Swin-transformer v2 \cite{DBLP:journals/corr/abs-2111-09883}. simCrossTrans\cite{simCrossTrans_cite} explored the cross modality transfer learning by using both the ConvNets and Vision Transfomers based on SUN RGB-D dataset based on the mask R-CNN \cite{maskrcnn} approach.\\

\textbf{Inter modality mixing: }
\cite{DBLP:journals/corr/abs-2203-01735} learns a dynamical and local linear interpolation between the different regions of cross-modality images in data-dependent fashion to mix up the RGB and infrared (IR) images. We explored both the static and dynamic mixing methods and found the static has a better performance. \cite{10.1145/3394171.3413821} uses an interpolation between the RGB and thermal images at pixel level. As we are training a unified model supporting both the single modality image and multiple modality images as input, we do not apply the interpolation to keep the original signal of each modality. We leverage the transformer architecture itself to automatically build up the gap between different modalities.\\

\textbf{Multimodal data fusion:}
Multimodal data fusion can be performed using three different approaches: early fusion, late fusion, and deep fusion. Early fusion combines various modalities of data at a lower-dimensional common space, and a feature extractor is then employed to extract relevant information. Early fusion has been applied to object detection and audio-visual processing, as demonstrated in \cite{early1} and \cite{early2}, respectively. Late fusion, on the other hand, employs independent feature extractors for different data sources and merges the extracted features in the final stage. Classical works on deep fusion for action recognition, gesture segmentation and recognition, and emotion recognition are demonstrated in \cite{late1}, \cite{late2}, and \cite{late3}, respectively. Deep fusion is characterized by fusing data at various stages of model training, transforming the input data into a higher-level representation through multiple layers, and allowing for the fusion of diverse modalities into a single shared representation layer. Various works such as \cite{zongwei1,zongwei2,zongwei3,deep1,deep2,DBLP:journals/corr/ChenMWLX16} have applied deep fusion to object detection. The study in \cite{early_late_deep} explores all three fusion methods for indoor semantic segmentation. In our research, we have chosen to adopt the early fusion approach for multimodal data processing.\\

\textbf{Transfer learning with same modality or cross modality:} Transfer learning is widely used in computer vision (CV), natural language processing (NLP) and biochemistry. Most transfer learning systems are based on the same modality (e.g. RGB image in CV and text in NLP). For the CV, common transfer learning is based on supervised way such as works in R-CNN \cite{DBLP:journals/corr/GirshickDDM13}, Fast RCNN \cite{DBLP:conf/iccv/Girshick15}, Faster RCNN \cite{DBLP:conf/nips/RenHGS15}, FPN \cite{Lin_2017_CVPR}, mask R-CNN \cite{maskrcnn}, YOLO\cite{DBLP:journals/corr/RedmonDGF15}, YOLO9000\cite{DBLP:journals/corr/RedmonF16}, RetinaNet \cite{DBLP:journals/corr/abs-1708-02002} use a pretrained backbone network model based on ImageNet classification task and the model is further trained based on the following task datasets such as COCO to achieve object detection or/and instance segmentation tasks. In the NLP, the transfer learning such as BERT \cite{devlin-etal-2019-bert}, GPT\cite{gpt1}, GPT-2\cite{gpt2}, GPT-3\cite{gpt3} are mainly based on self-supervised way and achieve great success. Inspired by the success of the self-supervised way transfer learning, the CV community is also exploring the self-supervised way to explore new possibilities, one recent work which is similar to the BERT in NLP is MAE \cite{DBLP:journals/corr/abs-2111-06377}. The MolGNN \cite{molgnn1,liu2021covid} in bioinformatics use a self-supervised way based on Graph Neural network (GNN) in the pretraining stage and achieve good performance in a few shot learning framework for the following subtasks. For this work, we explore the cross modality transfer learning from a pretrained model under the supervised learning approach. Recently, Frustum-Voxnet \cite{Shen_2020_WACV} used pretrained weights from the RGB images to fine tune the point cloud converted pseudo image based on ConvNets\cite{10.1162/neco.1989.1.4.541}. simCrossTrans\cite{simCrossTrans_cite} further explored the cross modality transfer learning by using both the ConvNets\cite{10.1162/neco.1989.1.4.541} and Vision Transfomers\cite{dosovitskiy2021an,liu2021Swin} and showed significant improvement.\\
\section{Methodology}
In this section, we will describe our approach for converting structured point clouds to pseudo images, the methods we use for mixing various modalities, as well as our detection frameworks.
\subsection {Convert point clouds to pseudo 2D image}
\begin{figure}[H]
\begin{center}
   \includegraphics[width=1.0\linewidth]{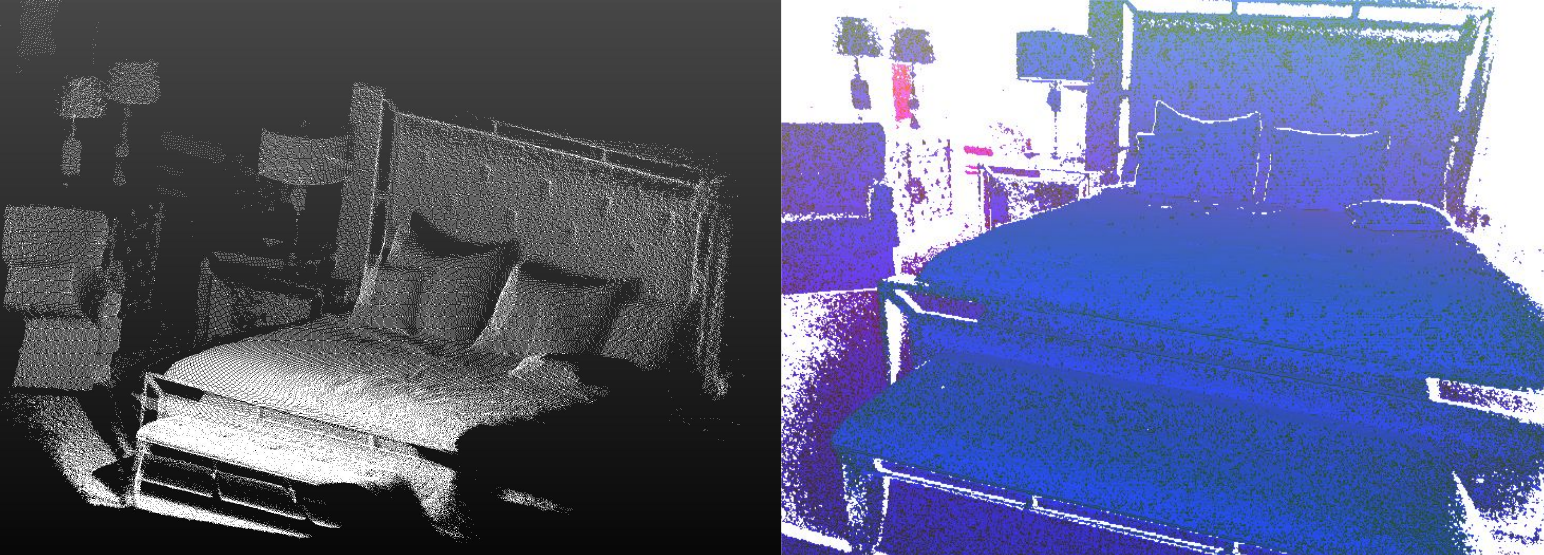}
\end{center}
    \caption{An example of converted pseudo three channel image from the point cloud. This example is one image selected from the validation set of the SUN RGB-D dataset. The corresponding RGB one can be found in Figure \ref{fig:main}}
\label{fig:pc2dhs}
\end{figure}
In order to use pretrained models based on RGB images, we convert point clouds to pseudo 2D images with 3 channels. The point clouds can be converted to HHA or any three channels from DHASM introduced in \cite{allancnn}.\\

For this work, we follow the same approaches in Frustum VoxNet\cite{Shen_2020_WACV} and simCrossTrans \cite{simCrossTrans_cite} by using DHS to project 3D depth data to 2D images \cite{allancnn}.
Here is a summary of the DHS encoding method. Similar to\cite{GuptaGAM14,allancnn}, we adopt \textbf{D}epth from the sensor and \textbf{H}eight along the sensor-up (vertical) direction as two reliable measures. \textbf{S}igned angle was introduced in \cite{OnlineClassificationStamos}. Let us denote as $X_{i,k} = [x_{ik},y_{ik},z_{ik}]$ the vector of 3D coordinates of the $k$-th point in the $i$-th scanline. Knowledge of the vertical direction (axis $\mathbf{z}$) is provided by many laser scanners, or even can be computed from the data in indoor or outdoor scenarios (based on line/plane detection or segmentation results from machine learning models) and is thus assumed known. Define $D_{i,k} = X_{i,k+1} - X_{i,k}$ (difference of two successive measurements in a given scanline $i$), and $A_{ik}$: the angle of the vector ${D_{i,k}}$ with the pre-determined $\mathbf{z}$ axis (0 to 180 degrees). The \textbf{S}igned angle $S_{ik} =sgn(\mathbf{D_{i,k}}\cdot \mathbf{D_{i,k-1}})*A_{ik}$: the sign of the dot product between the vectors $D_{i,k}$ and $D_{i,k-1}$, multiplied by $V_{ik}$. This sign is positive when the two vectors have the same orientation and negative otherwise. Those three channel pseudo images are normalized to 0 to 1 for each channel. Some samples DHS images can be seen in Figure \ref{fig:pc2dhs} and \ref{fig:aug}.
\subsection {Inter modality mixing}
\begin{figure}[H]
\begin{center}
   \includegraphics[width=0.9\linewidth]{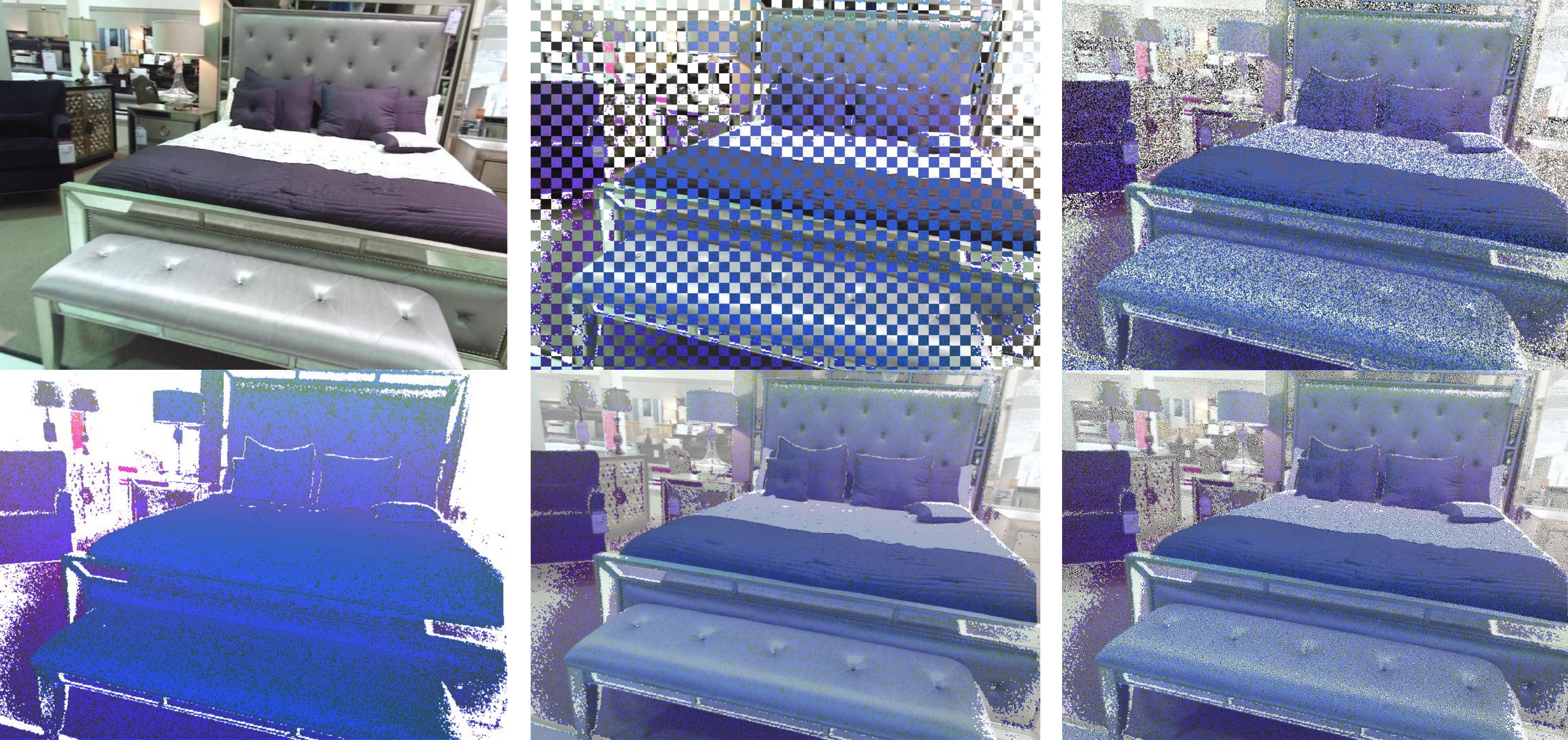}
\end{center}
    \caption{Inter modality mixing. Left column: original RGB image (up) and the original DHS image (bottom). Middle column: the Chessboard Per Patch Mixing image.  Patch size of 15 by 15 pixels (up) and 1 by 1 pixel (bottom). Right column: the Stochastic Flood Fill Mixing image. Edge connection probability of 0.5, 0.5 for RGB and DHS separately (up), probability of 0.1 and 0.1 for RGB and DHS separately (bottom).}
\label{fig:aug}
\end{figure}

In order to expand the input options for our unified model, we introduce an inter-modality mixing approach that enables us to combine images from different modalities into a three-channel image for consumption by the model. This approach allows us to enhance the model's capabilities without modifying its architecture. By training a model using RGB, DHS images, and the mixed RGB and DHS images, we can achieve a unified detector that is capable of processing different modalities as input.

Various techniques can be employed to fuse images from different modalities, and we propose two approaches:
\begin{itemize}
    \item Per Patch Mixing (PPM): divide the whole image into different patches with equal patch size. Randomly or alternatively select one image source for each patch.
    \item Stochastic Flood Fill Mixing (SFFM): Using a stochastic way to mix the images from different modalities.
\end{itemize}
We implement the Per Patch Mixing approach with relative simplicity. Specifically, for each patch in the image, we alternatively selected a modality image to assign to that patch. Moreover, we opted to utilize square patches for our implementation. As a result, the mask for selecting the modality image for each patch resembles a chessboard pattern, leading us to refer to our implementation as the Chessboard Per Patch Mixing (CPPM). Examples of the CPPM are shown in the middle of Figure \ref{fig:aug}.\\
\begin{figure}[H]
\begin{center}
   \includegraphics[width=0.9\linewidth]{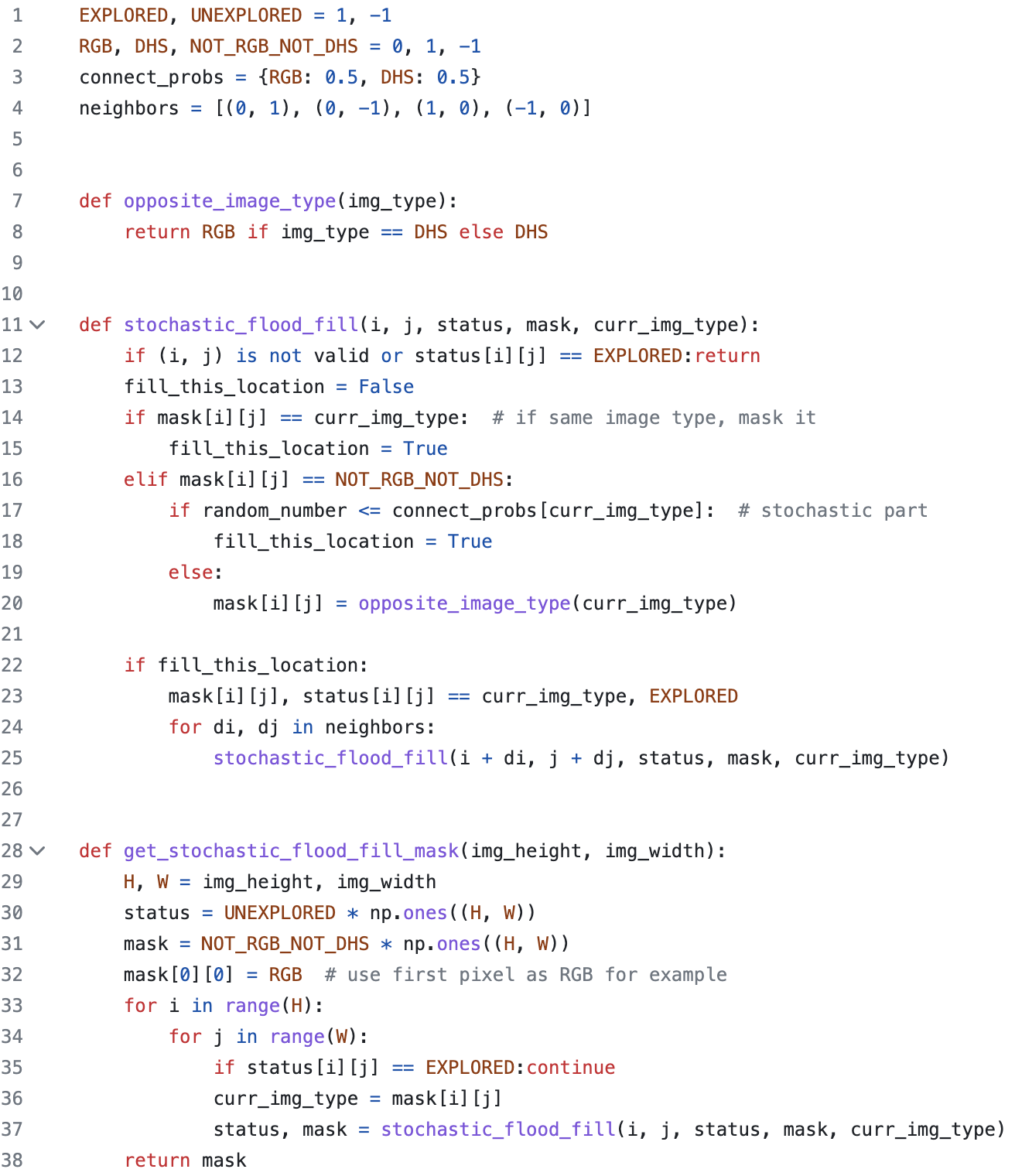}
\end{center}
    \caption{The pseudocode for the Stochastic Flood Fill Mixing is presented in Python style, with line 17 to 20 representing the stochastic aspect that differentiates it from the original flood fill algorithm. The mask is used to determine which image's pixel value should be used in generating the mixing image.}
\label{fig:sff}
\end{figure}
The Stochastic Flood Fill Mixing technique is an adaptation of the flood fill algorithm \cite{enwiki:floodfill}. The approach involves establishing connections between neighboring pixels with a probability $p$, with separate probabilities for the RGB and DHS modalities. The algorithm can be implemented using four or eight neighbors to build the graph, with the latter including additional diagonal offsets. In our experiments, we used the four neighbor approach. The Python-style pseudocode for this algorithm is illustrated in Figure \ref{fig:sff}, while examples of SFFM are shown on the right side of Figure \ref{fig:aug}.\\
\subsection{2D detection framework}
For the purpose of 2D detection and instance segmentation, we adopt the conventional object detection framework, namely Mask R-CNN \cite{maskrcnn} which is implemented in MMDetection \cite{mmdetection}. It follows a two-stage approach \cite{DBLP:journals/corr/abs-1905-12683}, namely region proposal and detection/segmentation, for accomplishing detection and segmentation tasks. During the fine-tuning of the model on SUN RGB-D dataset, we disable the training of the mask branch. However, even with the default weights from the pre-trained model, the mask prediction branch can still generate acceptable mask predictions, as demonstrated in Figure \ref{fig:main}. This observation aligns with the findings of the simCrossTrans \cite{simCrossTrans_cite} research.\\
\subsection{2D detection backbone networks}
For the backbone network, we use Swin Transformers\cite{liu2021Swin}, specifically we explored \textbf{Swin-T}iny and \textbf{Swin-S}mall's performance. The complexity of Swin-T and Swin-S are similar to those of ResNet-50 and ResNet-101, respectively. The window size is set to $M = 7$ by default. The query dimension of each head is $d = 32$, and the expansion layer of each MLP is $\alpha = 4$. The architecture hyper-parameters of these two models are:
\begin{itemize}
    \item Swin-T: C = 96, layer numbers = $\{2, 2, 6, 2\}$
    \item Swin-S: C = 96, layer numbers = $\{2, 2, 18, 2\}$
\end{itemize}
where C is the channel number of the hidden layers in the first stage. Detail of the model architecture, please check the Swin Transformers\cite{liu2021Swin} paper.\\
\subsection{SUN RGB-D dataset used in this work}
SUN RGB-D\cite{Song_2015_CVPR} dataset is an indoor dataset which provides both the point cloud and RGB images. In this work, since we are building a 3D only object detection system, we only use the point clouds for fine tuning. The RGB images are not used during the fine tuning process. For the point clouds, they are collected based on 4 types of sensors: Intel RealSense, Asus Xtion, Kinect v1 and Kinect v2. The first three sensors are using an IR light pattern. The Kinect v2 is based on time-of-flight. The longest distance captured by the sensors are around 3.5 to 4.5 meters.\\
SUN RGB-D dataset splits the data into a training set which contains 5285 images and a testing set which contains 5050 images. For the training set, it further splits into a training only, which contains 2666 images and a validation set, which contains 2619 images. Similar to \cite{DBLP:journals/corr/SongX15,Lahoud_2017_ICCV,Shen_2020_WACV,Shen_2020_v2} , we are fine-tuning our model based on the training only set and evaluate our system based on the validation set.\\
\subsection{Pre-training}
Both the Swin-T and Swin-S based networks\footnote{The pretrained weights are loaded from mmdetection \cite{mmdetection}.} are firstly pre-trained on ImageNet \cite{deng2009imagenet} and then pre-trained on the COCO dataset \cite{DBLP:journals/corr/LinMBHPRDZ14}.\\
\textbf{Data augmentation} When pre-training on COCO dataset, the image augmentations are applied during the training stage by: randomly horizontally flipping the image with probability of 0.5; randomly resizing the image with width of 1333 and height of several values from 480 to 800 (details see the configure file from the github repository); randomly cropping the original image with size of 384 (height) by 600 (width) and resizing the cropped image to width of 1333 and height of several values from 480 to 800.\\
\subsection{Fine-tuning}
\textbf{Data augmentation}: We follow the same augmentation with the pre-train stage. The raw input images have the width of 730 and height of 530. Those raw images are randomly resized and cropped during the training. During testing, the images are resized to width of 1120 and height of 800 which can be divided by 32.

\textbf{Hardware:} For the fine-tuning, we use a standard single NVIDIA Titan-X GPU, which has 12 GB memory. We fine-tune the network for 133K iterations for 100 epochs. It took about 29 hours for Swin-T based network with batch size of 2 (for 133K iterations) for the RGB only model. For the UODDM without the inter modality mixing, it took about 2 days to train the model. For with the inter modality mixing, the speed depends on the number of inter modality mixing images added to the training data. 

\textbf{Fine-tuning subtasks:} We focus on the 2D object detection performance, so we fine-tune the model based on the 2D detection related labels. Similar to simCrossTrans \cite{simCrossTrans_cite}, we kept the mask branch without training to further verify whether reasonable mask detection can be created by using the weights from the pre-train stage.
\section{Results on SUN RGB-D dataset}
\subsection{Experiments}
The primary focus of our experiments centers around the training of the model using diverse input data and the comparison of performance differences. Specifically, we first trained a unified model on both RGB and DHS images for the UODDM without inter-modality mixing. In contrast, for the UODDM with inter-modality mixing, we augmented the training data with inter-modality mixing images, in addition to the RGB and DHS images.
\subsection{Evaluation metrics} Following the previous works mentioned in Table \ref{2d_res}, we firstly use the AP50: Average Precision at IoU = 0.5 as evaluation metric. We also use the COCO object detection metric which is AP75: Average Precision at IoU = 0.75 and a more strict one: AP at IoU = .50:.05:.95 to evaluate the 2D detection performance. 
\subsection{Evaluation subgroups} 
We use the same subgroups as simCrossTrans\cite{simCrossTrans_cite} to evaluate the performance. The subgroups are SUNRGBD10, SUNRGBD16, SUNRGBD66 and  SUNRGBD79, which have 10, 16, 66 and 79 categories. Detail list of those sub groups can be found in simCrossTrans\cite{simCrossTrans_cite}.
\subsection{The performance of UODDM without the inter modality mixing}
We first evaluate the performance of UODDM without inter modality mixing. For this one, the model is trained based on both the RGB and DHS images. Our model architecture is the same as  the simCrossTrans \cite{simCrossTrans_cite} work, which is using only the DHS image to train the model. We train a RGB images only model based on the same network to compare with the UODDM one's performance. 
\begin{figure}[H]
\begin{center}
   \includegraphics[width=1.0\linewidth]{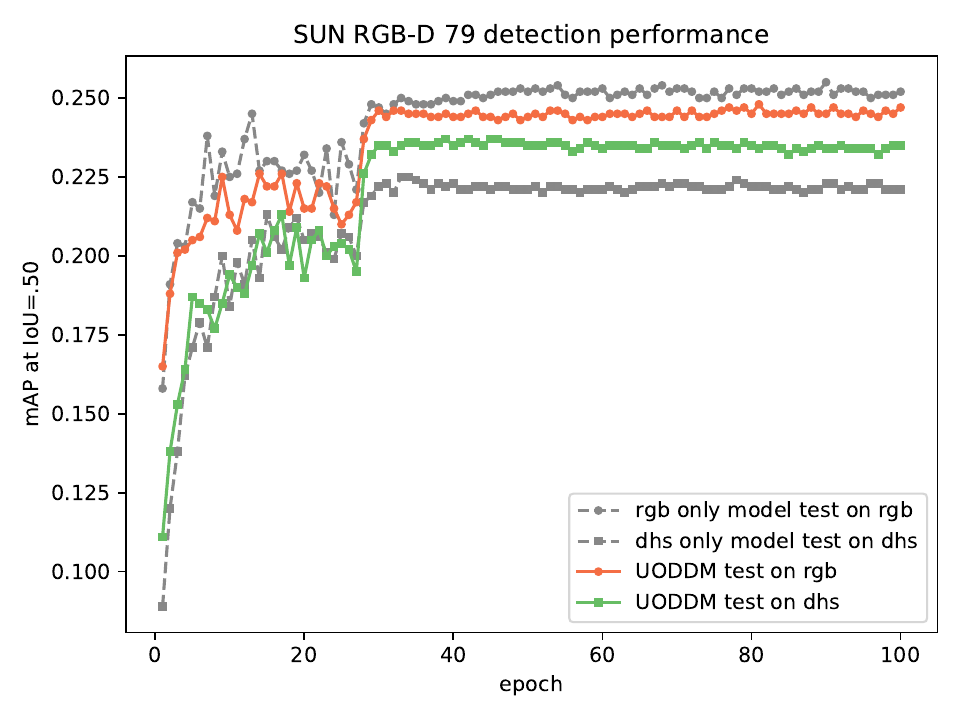}
\end{center}
    \caption{Comparison of the unified model and separate models performance with the training epochs. The RGB only one is our new trained model based on RGB images. The DHS only model is from simCrossTrans \cite{simCrossTrans_cite}. The UODDM is trained based on both the RGB and DHS images. The backbone for all those experiments are based on the Swin-T model.}
\label{fig:main_res}
\end{figure}
The performance evaluation of our proposed UODDM approach, measured in terms of mean average precision (mAP50), on the SUNRGBD79 dataset is presented in Figure \ref{fig:main_res}. The results show that the UODDM model performs exceptionally well on both RGB and DHS images. Additionally, it is evident that the UODDM model significantly outperforms the DHS-only model in terms of performance on DHS images, which can be attributed to the inter-modality transfer learning from the RGB images. However, this performance improvement on DHS images comes at a slight cost of performance reduction on RGB images. Nevertheless, the UODDM model's overall performance is promising as it is a single model that can handle different modalities, making it more efficient than maintaining two separate architectures or a single architecture with two different sets of weights. This efficiency is particularly valuable for robotics and edge devices, where a seamless perception system can be built, even when transitioning from daytime to nighttime scenarios. Table \ref{rgb_dhs} provides additional results for our UODDM and single-modality models, reinforcing the same conclusions.\\
\begin{table}[H]
\scriptsize
\begin{center}
\resizebox{1.0\linewidth}{!}{
\begin{tabular}{|c|c|c|c|c|c|c|}
\hline
model &	test on	& backbone&	sunrgbd10 &	sunrgbd16 &	sunrgbd66&	sunrgbd79\\
\hline \hline
RGB only (ours)	&	RGB&Swin-T&	\textbf{54.2}&	52.3&	\textbf{29.3}&	\textbf{25.2}\\
UODDM (ours)& RGB	&Swin-T&	53.9&	\textbf{52.5}&	28.7&	24.7\\
\hline \hline
DHS only (simCrossTrans \cite{simCrossTrans_cite})&DHS	&	Swin-T&	55.8&	52.7&	26.1&	22.1\\
UODDM (ours)& DHS&	Swin-T&	\textbf{56.6}&	\textbf{53.4}&	\textbf{27.7}&	\textbf{23.5}\\
\hline
\end{tabular}}
\end{center}
\caption {Result comparison based on mAP50 for different subgroups of UODDM and single modality only models.}
\label{rgb_dhs}
\end{table}
\subsection{The performance of UODDM with the inter modality mixing}
\begin{table}[H]
\scriptsize
\begin{center}
\resizebox{1.0\linewidth}{!}{
\begin{tabular}{|c|c|c|c|c|c|c|}
\hline
model &	test on	& backbone&	sunrgbd10 &	sunrgbd16 &	sunrgbd66&	sunrgbd79\\
\hline \hline
UODDM & RGB	&Swin-T&	53.9&	52.5&	\textbf{28.7}&	\textbf{24.7}\\
UODDM + SFFM & RGB	&Swin-T&24.6&   17.5&   19.2&   20.1	\\
UODDM + CPPM & RGB	&Swin-T&54.2&   51.9&   27.7&   23.7	\\
UODDM + CPPM & RGB	&Swin-S&\textbf{54.6}&   \textbf{52.7}&   27.5&   23.6	\\
\hline \hline
UODDM & DHS&	Swin-T&	56.6&	\textbf{53.4}&	\textbf{27.7}&	\textbf{23.5}\\
UODDM + SFFM & DHS	&Swin-T&25.6&   18.7&   20.0&   21.3	\\
UODDM + CPPM & DHS	&Swin-T&55.8&   52.8&   26.3&   22.4	\\
UODDM + CPPM & DHS	&Swin-S&\textbf{57.4}&   52.5&   24.8&   21.1	\\
\hline \hline
UODDM + CPPM & CPPM	&Swin-T&58.1&  55.8&   \textbf{29.5}&   \textbf{25.2}	\\
UODDM + CPPM & CPPM	&Swin-S&\textbf{58.4}&   \textbf{56.1}&   28.4&   24.5	\\
\hline
\end{tabular}}
\end{center}
\caption {Result comparison based on mAP50 for different subgroups of UODDM and single modality only models.}
\label{rgb_dhs_imm}
\end{table}
In our study, we investigated two different methods for inter-modality mixing, namely SFFM and CPPM. For SFFM, we generated six mixing images for each RGB and DHS image pair, with connection probabilities for RGB and DHS pixels being randomly selected from the range of 0.1 to 0.9. The first pixel's RGB and DHS masks were randomly initialized with equal probability. In contrast, for CPPM, we used square patches of size 1 by 1, resulting in one CPPM image for each RGB and DHS image pair. The performance of both approaches was evaluated and presented in Table \ref{rgb_dhs_imm}. Notably, the results suggest that the UODDM with CPPM outperforms the UODDM with SFFM. We attribute this to the generation of an excessive number of random images by SFFM, which can negatively impact the performance of the unified network on RGB and DHS images. Conversely, CPPM provides comparable performance to the plain UODDM model. Furthermore, the use of the CPPM image generated from both RGB and DHS images led to the best 2D detection performance. Given the ability of UODDM with CPPM to support RGB, DHS, and CPPM images from RGB and DHS, we propose it as a more powerful unified model.
\subsection {Influence of different backbone networks}
Table \ref{rgb_dhs_imm} presents the results obtained by using Swin-T and Swin-S as the backbone networks. It is observed that Swin-S is a more powerful network; however, the performance gain achieved is limited. Therefore, we propose the usage of the lightweight Swin-T as the backbone network to achieve a faster inference speed, as depicted in Table \ref{infer_time_comp_}.
\subsection{Comparisons with other methods}
\begin{table}
\begin{center}
	\resizebox{\textwidth}{!}{
\begin{tabular}{|c|l|l|c|c|c|c|c|c|c|c|c|c|c|c|c|c|c|c|c|c|}
\hline
Image Source &Methods	&Backbone& bed     &toilet&  \shortstack{night \\stand}&     bathtub&        chair  &dresser &sofa& table & desk&	bookshelf	&\shortstack{sofa\\ chair}&\shortstack{kitchen\\ counter}&\shortstack{kitchen\\ cabinet}&\shortstack{garbage\\ bin}&microwave&sink&\shortstack{SUNRGBD10\\$mAP_{50}$}&\shortstack{SUNRGBD16\\$mAP_{50}$}\\
\hline\hline
  \multirow{4}{*}{RGB}&      2D-driven\cite{Lahoud_2017_ICCV}&VGG-16&74.5&  86.2& 49.5&   45.5&   53.0&29.4&49.0&42.3&	22.3&	45.7&	N/A&N/A&N/A&N/A&N/A&N/A&49.7&N/A\\
&  	Frustum PointNets\cite{Qi_2018_CVPR}&adjusted VGG from SSD&56.7&   43.5& 37.2&  \textbf{81.3}&   64.1&\textbf{33.3}&57.4&\textbf{49.9}&	\textbf{77.8}&	\textbf{67.2}&	N/A&N/A&N/A&N/A&N/A&N/A&56.8&N/A\\
&	F-VoxNet\cite{Shen_2020_WACV}&ResNet 101	&81.0&  89.5& 35.1&   50.0&   52.4&21.9&53.1&37.7	&18.3	&40.4	&47.8&22.0&29.8&52.8&39.7&31.0&47.9&N/A\\
&rgb only model (ours) &Swin-T&  83.2&   \textbf{93.9}&   51.8&   54.2&   60.4&   23.7&   51.3&   46.3&   22.5&   54.4&   60.4&   \textbf{32.7}&   39.8&   67.0&   48.1&   47.3&   54.2&   52.3\\
&UODDM (ours)  &Swin-T&  83.6&   87.1&   53.3&   58.8&   62.5&   22.6&   54.2&   46.8&   22.0&   48.0&   63.7&   28.9&   38.4&   67.1&   57.4&   46.2&   53.9&   52.5\\
&UODDM + CPPM (ours)  &Swin-T&83.6&   88.6&   53.0&   59.1&   60.8&   26.5&   50.7&   46.1&   22.0&   52.0&   60.6&   31.2&   37.8&   64.7&   54.3&   40.1&   54.2&   51.9\\
\hline\hline
\multirow{4}{*}{Point Cloud only}& 	F-VoxNet\cite{Shen_2020_WACV}&ResNet 101&78.7 & 77.6& 34.2&   51.9  &51.8&16.5&48.5	&34.9	&14.2	&19.2 &48.7&19.1&18.5&30.3&22.2&30.1&42.8&37.3\\
&	simCrossTrans\cite{simCrossTrans_cite}& Swin-T&87.2&	87.7&	51.6	&69.5	&69.0	&27.0	&60.5	&48.1&	19.3&	38.3&	68.1&	30.7&	35.5&	61.2&41.9&	47.7&55.8&52.7\\
&UODDM (ours)   &Swin-T&  \textbf{88.1}&   87.6&   53.8&   66.8&   \textbf{69.5}&   28.7&   \textbf{62.2}&   47.2&   19.7&   41.9&   \textbf{68.5}&   28.5&   35.8&   62.8&   41.9&   \textbf{51.5}&   56.6&   53.4\\
&UODDM + CPPM (ours)  &Swin-T&88.0&   85.6&   51.8&   68.3&   68.6&   26.9&   61.6&   45.5&   20.2&   41.7&   67.6&   29.2&   33.0&   61.6&   47.4&   47.5&   55.8&   52.8\\
\hline\hline
\multirow{2}{*}{\shortstack{RGB $\&$\\Depth/Point Cloud}}&RGB-D RCNN\cite{GuptaGAM14}&VGG&76.0&   69.8&37.1&   49.6&   41.2   &31.3&42.2&43.0& 16.6&   34.9 &N/A&N/A&N/A&46.8&N/A&41.9&44.2&N/A\\
&UODDM 
+ CPPM (ours)&Swin-T&86.5&   91.0&   \textbf{54.4}&   70.2&   67.2&   30.3&   57.5&   48.7&   22.8&   52.7&   66.6&   29.2&   \textbf{41.2}&   \textbf{68.6}&   \textbf{57.9}&   48.0&  \textbf{ 58.1}&   \textbf{55.8}  \\
\hline
\end{tabular}}
\end{center}
	\caption{2D detection results based on SUN RGB-D validation set. Evaluation metric is average precision with 2D IoU threshold of 0.5.}
	\label{2d_res}
\end{table}
In Table \ref{2d_res}, we present a detailed comparison of per category results with previous works. Specifically, we evaluate the performance of our approach under three different input scenarios: RGB image, point cloud, and a combination of RGB and point cloud data using our proposed inter modality mixing method.\\

When considering RGB image as input, we observe that our best performing UODDM with CPPM or RGB-only model achieve slightly worse performance (54.2 mAP50 on SUNRGBD10) than the state-of-the-art Frustum PointNets \cite{Qi_2018_CVPR}. On the other hand, when utilizing only point cloud as input, our plain UODDM model (without inter modality mixing) demonstrates a slightly better performance (56.6 mAP50 on SUNRGBD10) compared to the previous state-of-the-art \cite{simCrossTrans_cite}.\\

Remarkably, our proposed UODDM with CPPM significantly outperforms the previous best results obtained by RGB-D RCNN \cite{GuptaGAM14} (58.1 mAP50 on SUNRGBD10), in the scenario where both RGB and point cloud data are available. Notably, most prior works have focused on utilizing either RGB or point cloud data, with limited exploration of mixing methods for these modalities. Therefore, the proposed inter modality mixing method constitutes a significant contribution to the field.\\

Moreover, our UODDM with CPPM method demonstrates a substantial performance gain compared to the strongest 2D detector from RGB images, i.e., Frustum PointNets \cite{Qi_2018_CVPR}. Specifically, our approach achieves 58.1 mAP50 on SUNRGBD10, which is superior to the performance of Frustum PointNets (56.8 mAP50 on SUNRGBD10).
\subsection{More results based on extra evaluation metrics}
More results based on mAP/mAP75 can be found in the appendix.
\subsection{Number of parameters and inference time}
\begin{table}[H]
\scriptsize
\begin{center}
               \resizebox{1.0\linewidth}{!}{
\begin{tabular}{|c|c|c|c|c|c|}
\hline
      Method & Backbone Network& \# Parameters (M) & GFLOPs &Inference Time (ms) & FPS\\
\hline\hline
      F-VoxNet \cite{Shen_2020_WACV} & ResNet-101 & 64 &-& 110 & 9.1\\
      \hline
      simCrossTrans\cite{simCrossTrans_cite} & ResNet-50 & 44 & 472.1&70 & 14.3\\
       simCrossTrans \cite{simCrossTrans_cite}& Swin-T & 48 & 476.5&105 & 9.5\\
        UODDM (ours) & Swin-T & 48 & 476.5&105 & 9.5\\
       UODDM (ours) & Swin-S  &69 &419.7 &148& 6.8\\
\hline
\end{tabular}}
\end{center}
\caption {Number of parameters and inference time comparison. All speed testing are based on a standard single NVIDIA Titan-X GPU.}
\label{infer_time_comp_}
\end{table}
Table \ref{infer_time_comp_} presents the number of parameters and inference time for our proposed network architecture. The inference time reported for the Swin-T based network is the same as that reported in the simCrossTrans \cite{simCrossTrans_cite} paper, as we used the same network and hardware. However, since the Swin-S based network is larger, the inference time is slower, which is expected.

\section{Conclusion}
This paper presents a novel unified model capable of processing various types of data modalities, including RGB camera, DHS from depth sensor, and inter-modality mixing images from both RGB and DHS sources. The proposed system is highly versatile and exhibits exceptional performance in different scenarios, where the availability of the sensors may vary. The use of RGB camera during the day and depth sensor at night results in a more eco-friendly and sustainable solution, which also benefits from the improved performance achieved with the inter-modality mixing technique. Our results demonstrate the superior performance of the proposed unified model compared to single-modality models, making it an efficient and powerful solution for various practical applications.
\section{Acknowledgement}
We would like to express our gratitude to Zhujun Li and Jaime Canizales for their valuable comments and advice during the development of this work. We would also like to thank Zhujun Li for suggesting the name ``chessboard" to describe the method of alternatively selecting a modality image based on square patch.\\

\bibliographystyle{splncs}
\bibliography{egbib}

\section{More results}
\begin{table}[H]
\scriptsize
\begin{center}
               \resizebox{1.0\linewidth}{!}{
\begin{tabular}{|c|c|c|ccc|ccc|ccc|ccc|ccc|}
\hline
     \multirow{2}{*}{Method}  &\multirow{2}{*}{Test on}& \multirow{2}{*}{Backbone Network}& \multicolumn{3}{c|}{SUNRGBD10}&\multicolumn{3}{c|}{SUNRGBD16}&\multicolumn{3}{c|}{SUNRGBD66}&\multicolumn{6}{c|}{SUNRGBD79}\\
    && &AP&AP$_{50}$&AP$_{75}$&AP&AP$_{50}$&AP$_{75}$&AP&AP$_{50}$&AP$_{75}$&AP&AP$_{50}$&AP$_{75}$&AP$_S$&AP$_M$&AP$_L$\\
\hline\hline
     rgb only&       rgb     &Swin-T&        29.6&   54.2&   28.9&   28.6&   52.3&   28.3&   15.4&   29.3&   14.3&   13.1&   25.2&   12.1&   1.0&    5.2&    16.8\\
UODDM   &rgb&   Swin-T& 30.7&   53.9&   30.9&   29.5&   52.5&   29.7&   15.3&   28.7&   14.5&   13.1&   24.7&   12.2&   0.2&    4.6&    16.9\\
UODDM + CPPM&   rgb&    Swin-T & 30.9&   54.2&   30.5&   29.2&   51.9&   28.9&   14.5&   27.7&   13.2&   12.4&   23.7&   11.1&   0.7&    4.1&    15.7\\
UODDM&  rgb&    Swin-S  &31.8&  54.7&   32.3&   30.0&   52.5&   29.6&   15.1&   28.2&   13.7&   12.9&   24.4&   11.6&   0.4&    4.2&    16.1\\
UODDM + CPPM&   rgb&    Swin-S&  31.3&   54.6&   31.3&   29.9&   52.7&   29.6&   14.6&   27.5&   13.5&   12.4&   23.6&   11.4&   0.7&    3.9&    15.5\\
\hline\hline
simCrossTrans\cite{simCrossTrans_cite}&  dhs&    Swin-T & 33.3&   55.8&   34.7&   30.7&   52.7&   31.5&   14.3&   26.1&   14.0&   12.0&   22.1&   11.7&   0.6&    4.7&    15.2\\
UODDM&  dhs&    Swin-T& 34.0&   56.6&   34.9&   31.4&   53.4&   31.9&   15.4&   27.7&   14.8&   13.0&   23.5&   12.4&   0.4&    4.6&    16.4\\
UODDM + CPPM&   dhs     &Swin-T&        33.8&   55.8&   35.9&   31.3&   52.8&   32.5&   14.9&   26.3&   14.7&   12.6&   22.4&   12.4&   0.6&    4.4&    16.0\\
UODDM&  dhs&    Swin-S  &\textbf{34.8}&  57.6&   \textbf{37.0}&   31.7&   53.7&   32.4&   14.8&   26.6&   14.2&   12.5&   22.6&   12.0&   0.7&    4.2&    15.7\\
UODDM + CPPM&   dhs&    Swin-S& 34.1&   57.4&   35.8&   30.9&   52.5&   31.8&   14.1&   24.8&   14.0&   11.9&   21.1&   11.8&   1.1&    4.5&    15.1\\
\hline\hline
UODDM + CPPM&   CPPM image&     Swin-T& 34.2&   58.1&   35.0&   32.6&   55.8&   33.2&   \textbf{16.3}&   \textbf{29.5}&   \textbf{15.9}&   \textbf{13.8}&   \textbf{25.2}&   \textbf{13.4}&   0.4&    5.0&    \textbf{17.1}\\
UODDM + CPPM&   CPPM image&     Swin-S& 34.6&   \textbf{58.4}&   36.0&   \textbf{33.0}&  \textbf{ 56.1}&   \textbf{34.4}&   15.9&   28.4&   15.7&   13.7&   24.5&   13.4&   \textbf{1.2}&   \textbf{ 5.4}&    16.8\\
\hline
\end{tabular}}
\end{center}
\caption {More results comparison based on AP@IoU = .75, AP and AP of different scales.}
\label{more_2d_res}
\end{table}
Besides the AP50, which was mainly used in previous works, we also use AP75 and AP to compare the results based on different methods. Meanwhile, we also report AP Across Scales of small, medium and large by following the same standard of COCO dataset. Those results can be found in Table \ref{more_2d_res}. From the results, we see that in general the UODDM with the CPPM can achieve the best performance on the CPPM image. This is mainly due to the fact that both the RGB and DHS images are used for the system. When only using the RGB image and only using the DHS image, the unified model UODDM with CPPM has similar performance as the single modality based model.\\
\end{document}